%% file: main.tex
\documentclass[sigconf]{acmart}

\AtBeginDocument{%
  }
\setcopyright{acmlicensed}
\copyrightyear{2018}
\acmYear{2018}
\acmDOI{XXXXXXX.XXXXXXX}
\acmConference[Conference acronym 'XX]{Make sure to enter the correct
  conference title from your rights confirmation email}{June 03--05,
  2018}{Woodstock, NY}
\acmISBN{978-1-4503-XXXX-X/2018/06}
\usepackage{multirow}
\usepackage{xspace}
\usepackage{url}
\usepackage{subcaption}
\usepackage{url}
\usepackage{multirow}
\usepackage{graphicx}
\usepackage{stfloats}
\usepackage{ulem}
\usepackage{bm}
\usepackage{tcolorbox}
\usepackage{listings}
\usepackage{tcolorbox}
\usepackage{wrapfig}
\usepackage{algorithm}
\usepackage{algpseudocode}
\usepackage{lipsum}
\usepackage{enumitem}
\usepackage{fancyvrb}
\usepackage{caption}
\usepackage{xcolor}
\usepackage{xcolor}

\definecolor{darkgreen}{RGB}{0,100,0}
\definecolor{darkred}{RGB}{139,0,0}
\newcommand{\method}{\textsc{ReC4TS}\xspace}
\newcommand{\data}{\textsc{Time-Thinking}\xspace}
\newcommand{\better}{\textcolor{darkgreen}}
\newcommand{\worse}{\textcolor{darkred}}
\newcommand{\observationbox}[2]{
    \begin{tcolorbox}[
        colback=green!2!white,
        colframe=green!20!gray,
        title=#1,
        fonttitle=\bfseries,
        coltitle=white,
        boxrule=0.5pt
    ]
    #2
    \end{tcolorbox}
}
\newcommand{\rejectionbox}[2]{
    \begin{tcolorbox}[
        colback=red!2!white,
        colframe=red!20!gray,
        title=#1,
        fonttitle=\bfseries,
        coltitle=white,
        boxrule=0.5pt
    ]
    #2
    \end{tcolorbox}
}

\begin{document}

\title{Evaluating System 1 vs. 2 Reasoning Approaches for Zero-Shot Time Series Forecasting: A Benchmark and Insights}

\author{Haoxin Liu,\xspace Zhiyuan Zhao, Shiduo Li, B. Aditya Prakash}
\email{{hliu763, leozhao1997, sli999, badityap}@gatech.edu}
\affiliation{\institution{Georgia Institute of Technology}
\city{Atlanta}
\state{GA}
\country{USA}
}

\begin{abstract}

Reasoning ability is crucial for solving challenging tasks. With the advancement of foundation models, such as the emergence of large language models (LLMs), a wide range of reasoning strategies has been proposed, including test-time enhancements, such as Chain-of-Thought, and post-training optimizations, as used in DeepSeek-R1. While these reasoning strategies have demonstrated effectiveness across various challenging language or vision tasks, their applicability and impact on time-series forecasting (TSF), particularly the challenging zero-shot TSF, remain largely unexplored. In particular, it is unclear whether zero-shot TSF benefits from  reasoning and, if so, what types of reasoning strategies are most effective.

To bridge this gap, we propose \method, the first benchmark that systematically evaluates the effectiveness of popular reasoning strategies when applied to zero-shot TSF tasks. \method conducts comprehensive evaluations across datasets spanning eight domains, covering both unimodal and multimodal with short-term and long-term forecasting tasks. More importantly, \method provides key insights: (1) Self-consistency emerges as the most effective test-time reasoning strategy; (2) Group-relative policy optimization emerges as a more suitable approach for incentivizing reasoning ability during post-training; (3) Multimodal TSF benefits more from reasoning strategies compared to unimodal TSF.  
Beyond these insights, \method establishes two pioneering starting blocks to support future zero-shot TSF reasoning research: (1) A novel dataset, \data, containing forecasting samples annotated with reasoning trajectories from multiple advanced LLMs, and (2) A new and simple test-time scaling-law validated on foundational TSF models enabled by self-consistency reasoning strategy.  All data and code are publicly accessible at: \url{https://github.com/AdityaLab/OpenTimeR}
\end{abstract}


\maketitle

\input{Section/Intro}

\input{Section/Related}
\input{Section/Settup}

\input{Section/New_Exp_1}
\input{Section/New_Exp_2}
\input{Section/New_Exp_3}
\input{Section/Toolkits}
\input{Section/Conclusion}

\bibliographystyle{ACM-Reference-Format}
\bibliography{reference}

\appendix
\input{Section/Appendix}

\end{document}

%% file: Section/Intro.tex
% \vspace{-0.05in}
\section{Introduction}\label{sec:intro}

Reasoning capabilities are fundamental to solving challenging tasks and have been the focus of research for decades~\cite{gebhardt1997reasoning, wustenberg2012complex}. Recent advancements in foundation models, particularly the rise of large language models (LLMs), have brought reasoning strategy research into the era of artificial intelligence~\cite{wei2022chain, wang2022self, pan2023automatically}. By integrating designed reasoning strategies with powerful foundation models, these models have demonstrated remarkable success in tackling complex real-world challenges, including math, planning, and engineering~\cite{bai2023qwen, song2023llm, sun2023adaplanner, jin2024llms}. Despite these advancements in extensive domains, the application of reasoning strategies in zero-shot time-series forecasting (TSF) remains largely unexplored. While recent time-series foundation models have initially demonstrated the feasibility of zero-shot TSF~\cite{das2024decoder, goswami2024moment, ansari2024chronos, shi2024time}, they primarily rely on the memory of large-scale time series data from the pre-training phase rather than employing explicit reasoning for zero-shot TSF.

However, reasoning is not only natural but also essential for zero-shot TSF. On the one hand, time-series data inherently contains rich causal relationships, making it a natural domain for reasoning-based approaches~\cite{moraffah2021causal, runge2023causal}. For instance, in epidemiological forecasting, the outbreak of a pandemic naturally reasons a subsequent increase in infections—an essential target for flu prediction~\cite{mathis2024evaluation}. On the other hand, and more critically, reasoning plays an important role in zero-shot TSF. Unlike conventional supervised TSF, which ``memorizes'' underlying correlation structures from training data, zero-shot TSF—designed as the primary objective for most time-series foundation models—depends on reasoning capabilities to ``interpret'' the pattern correlations within limited historical data, which have never been ``memorized,'' in order to infer future values. Nevertheless, existing time-series foundation models have yet to fully exploit them by incorporating effective reasoning strategies.

Therefore, a natural twofold research question (RQ) for reasoning strategies on zero-shot TSF arises: \textbf{``RQ1: Can zero-shot TSF benefit from enhanced reasoning ability?''} While reasoning is an integral part of time-series analysis, such as used in feature selection and statistical analysis~\cite{sun2015using, chen2004analyzing}, its effectiveness has yet to be systematically validated in zero-shot TSF with foundation models. Furthermore, if reasoning is indeed beneficial (as demonstrated later in our work), a subsequent question emerges: \textbf{``RQ2: What kind of reasoning strategies does zero-shot TSF need?''}  

Existing reasoning strategies are often categorized into two cognitive systems: Reasoning enhanced System 1, which enables fast and direct thinking, and System 2, which involves deep and analytical reasoning~\cite{kahneman2011thinking} built-in the foundation model. These two approaches differ distinctly in their implementation. Reasoning enhanced System 1 typically occurs during the inference stage, such as Chain-of-Thought (CoT) reasoning, possibly being further enhanced through self-consistency and self-correction~\cite{wei2022chain, wang2022self, pan2023automatically, kumar2024training}. In contrast, System 2 reasoning asks the model itself to promote deeper, structured thinking by incorporating incentives and rewards during post-training~\cite{ouyang2022training, guo2025deepseek}. Given the fundamental differences between System 1 and System 2 reasoning strategies in foundation models, addressing \textbf{RQ2} requires an in-depth investigation into their effectiveness for time-series forecasting. However, such studies remain unexplored, and a systematic benchmark to assess the impact of different reasoning strategies in TSF is still lacking. 

To bridge this gap and address the proposed twofold research questions, we introduce \method, a benchmark designed to systematically evaluate the effectiveness of various popular reasoning strategies in zero-shot TSF. To the best of our knowledge, this is the first benchmark study to investigate these questions. Our main contributions are summarized as follows:  
\begin{itemize}
    \item \textbf{New Research Direction:} We introduce a novel research direction focused on reasoning strategies for zero-shot TSF.  This research aims to understand when, what, and how reasoning strategies impact zero-shot TSF tasks.  
    \item \textbf{Comprehensive Benchmark:} We systematically evaluate the effectiveness of existing reasoning strategies for zero-shot TSF, conducting about 1,500 experiments. Our analysis covers both reasoning-enhanced System 1 and System 2 reasoning strategies across six foundation models.  
    \item \textbf{In-depth Insights:}  Our benchmark provides in-depth insights, revealing that self-consistency is the current most effective reasoning-enhanced System 1 strategy, and group-relative policy optimization is the current most suitable System 2 approach. Furthermore, we show that reasoning strategies are generally more effective in multimodal zero-shot TSF than unimodal zero-shot TSF. 
     \item \textbf{Open-Source Toolkits:} We release three key toolkits, including the evaluation suite, scaling law and reasoning data to comprehensively initiate future research on reasoning models for TSF. Specifically, we fully release \method as an easy-to-use evaluation suite, including the code, documents and experiment logs. we validate a new and simple test-time scaling law for zero-shot TSF with foundation time-series models enabled by self-consistency reasoning. Additionally, we introduce \data, the first TSF dataset with reasoning trajectories. 
\end{itemize}

%% file: Section/Related.tex
\vspace{-1mm}
\section{Related Work}

\par\noindent\textbf{(Zero-Shot) Time-Series Forecasting.} Time-series forecasting (TSF) is a fundamental task in time-series analysis with broad real-world applications, including economics, urban computing, and epidemiology~\cite{sezer2020financial, tabassum2021actionable, rodriguez2024machine}. The introduction of transformer-based architectures~\cite{vaswani2017attention} has significantly advanced TSF performance, setting new benchmarks in forecasting accuracy~\cite{zhou2021informer, zhou2022fedformer, Yuqietal-2023-PatchTST, liu2023itransformer}.  Building upon these advancements, recent research has explored foundation TSF models, which achieve competitive zero-shot forecasting performance comparable to supervised TSF models~\cite{das2024decoder, ansari2024chronos, goswami2024moment, shi2024time, ekambaram2025tiny, kamarthi2023large}. Trained on billions of data samples, these foundation models can generate accurate forecasts across diverse time series without requiring additional post-training or fine-tuning.

\par\noindent\textbf{Foundation Model Reasoning.} With the rise of foundation models, reasoning has become a critical research direction, enabling models to move beyond surface-level pattern recognition toward structured decision-making~\cite{wei2022chain, brown2020language}. Reasoning in foundation models primarily follows two paradigms: reasoning-enhanced System 1, which improves inference during test time for fast and intuitive thinking, such as Chain-of-Thought reasoning~\cite{wei2022chain, wang2022self, pan2023automatically, kumar2024training}, and System 2, which enables deep, analytical reasoning as an inherent capability of the model. System 2 reasoning is typically developed through post-training techniques such as reinforcement learning from human feedback (RLHF), reward shaping, knowledge distillation, and the more recent group-relative policy optimization (GRPO) in DeepSeek-R1~\cite{bai2022training, kwon2023reward, gou2021knowledge, xu2024survey, guo2025deepseek}, which allows foundation models to process more structured, deeper, and analytical thinking. Leveraging these advanced reasoning strategies, foundation models have demonstrated strong capabilities in handling complex tasks, including mathematical problem-solving, automated planning, and applications in engineering and science~\cite{bai2023qwen, zhou2024isr, sun2023adaplanner, song2023llm, guo2023can, goh2024large}.  

\par\noindent\textbf{Time-Series Reasoning.} Reasoning over time-series data is an important topic in time-series analysis~\cite{kauppinen2007modeling}. Conventional time-series reasoning primarily focuses on causal analysis, including causal discovery, feature selection, and graph-based analysis~\cite{eichler2012causal, sun2015using, runge2023causal, chen2004analyzing}. However, these conventional methods focus only on numerical modalities, making validation challenging and often limiting their application to synthetic or simplified scenarios. Recent advancements have broadened time-series reasoning beyond numerical data, incorporating multimodal perspectives. Notably, emerging research has explored reasoning over time-series data using LLMs~\cite{liu2024lstprompt, yan2025position} and vision-based approaches~\cite{liu2024picture}, expanding its scope and applicability. While these studies show promising potential for time series reasoning with advanced foundation models, they do not explicitly establish a connection between reasoning strategies and TSF tasks.

Importantly, existing studies and benchmarks focus on either (zero-shot) TSF ~\cite{aksu2024gift, wang2024deep, du2024tsi} or reasoning strategies~\cite{lin2024criticbench, hao2024llm, zhang2025benchmark}, no existing work systematically evaluates how different reasoning strategies impact zero-shot TSF. Our study bridges this gap and introduces a multimodal reasoning enhancement setup beyond conventional unimodal approaches. This multimodal setup incorporates textual context alongside numerical data, providing more comprehensive information for real-world forecasting, and setting a higher standardization for reasoning ability in zero-shot TSF tasks.

\begin{figure*}[t]
    \centering
    \includegraphics[width=\textwidth]{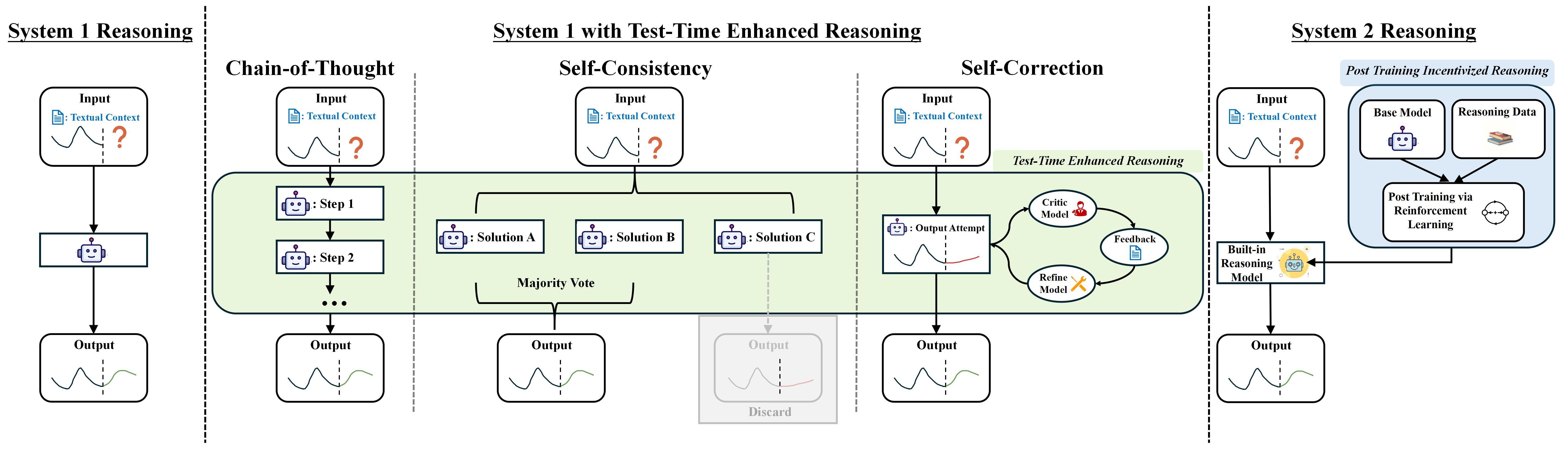}  % 或 tsf_comparison.pdf
    \vspace{-3mm}
    \caption{The reasoning strategies included in the proposed \method benchmark. \method systematically includes three mainstream approaches: the direct System 1, i.e., directly using generative models such as GPT-4o for reasoning; the test-time-enhanced System 1, including Chain-of-Thought, Self-Consistency, and Self-Correction; the post-training-empowered System 2, which enables built-in reasoning capabilities through reinforcement learning, such as DeepSeek-R1 \cite{guo2025deepseek}.}
    \label{fig:reason_strategy}
\end{figure*}

%% file: Section/Settup.tex
% \vspace{-1mm}
\section{\method: A Suite of Evaluating Reasoning Strategies for Zero-Shot TSF }
\method consists of four core modules: Datasets, Reasoning Strategies, Models, and Evaluations. We introduce these modules one by one in this section. More details are provided in Section~???\ref{sec:detail_bench}.
\subsection{Dataset module} The dataset module includes datasets from eight different domains, all containing both numerical time series and aligned textual context series, providing unified data support for downstream time series forecasting. As detailed in Table~\ref{tab:series}, these verified datasets~\cite{liu2025time,lin2024decoding} cover key domains such as Agriculture, Climate, Economy, Energy, Health, Security, Employment, and Traffic, with weekly and monthly frequencies. The textual context series consists of keyword-based web summaries, aligned by date with the numerical series. As shown in Figure~\ref{fig:OT_visualize}, these datasets exhibit diverse characteristics, enabling comprehensive evaluation. 
\subsection{Reasoning Strategies Module} As shown in Figure~\ref{fig:reason_strategy}, \method systematically includes three mainstream approaches for reasoning, following existing works \cite{pan2023automatically,plaat2024reasoning,xu2025towards}: (1) Direct System 1 Reasoning – directly using generative models such as GPT-4o for reasoning. (2) Test-Time-Enhanced System 1 Reasoning – incorporating techniques such as Chain-of-Thought \cite{wei2022chain}, Self-Consistency \cite{wang2022self}, and Self-Correction \cite{madaan2023self}. These approaches improve reasoning beyond System 1’s intuitive responses by performing additional computations during inference without modifying the model’s pre-trained weights. (3) Post-Training-Enabled System 2 Reasoning – such as DeepSeek-R1 \cite{guo2025deepseek}. Unlike test-time-enhanced System 1, System 2 reasoning is typically achieved through reinforcement learning and reasoning data during the post-training phase. System 2 has built-in reasoning capabilities, which are typically characterized by automatic long-chain thinking.

Specifically, Chain-of-Thought (CoT) prompts the model to break down complex problems into a series of logical steps before providing a final answer, mimicking human reasoning processes. The Self-Consistency method further enhances reasoning diversity by generating multiple reasoning paths in parallel and selecting the most consistent result. In contrast, the Self-Correction approach iteratively refines the model’s output through feedback, aiming to improve overall accuracy and reliability. For the System 2 strategies, also known as large reasoning models \cite{xu2025towards}, \method includes the closed-source o1-mini from OpenAI\footnote{\url{https://openai.com/index/openai-o1-mini-advancing-cost-efficient-reasoning/}} and Gemini-2.0-flash-thinking from Google\footnote{\url{https://cloud.google.com/vertex-ai/generative-ai/docs/thinking}}, as well as the open-source DeepSeek-R1 from DeepSeek. Compared to o1-Mini, which employs Proximal Policy Optimization (PPO) \cite{schulman2017proximal} by training two models simultaneously—a policy model for generating responses and a critic model for evaluating them—DeepSeek-R1 adopts Group Relative Policy Optimization (GRPO) \cite{shao2024deepseekmath}, which eliminates the need for a separate critic model by ranking multiple responses at once.
\subsection{Models Module} 

\method covers three series of foundational models, including both closed- and open-source models. Each series provides System 1 and System 2 versions. \uline{Note that since reasoning strategies for foundational time-series models have not yet been studied and are difficult to implement directly, reusing foundational language models for zero-shot TSF—which have been widely validated by existing works—is currently the best choice}~\cite{xue2023promptcast,gruver2023large,liu2024lstprompt,jintime,caotempo}.

Specifically, \method includes OpenAI's GPT-4o and o1-mini, Google's Gemini-2.0-Flash and Gemini-2.0-Flash-Thinking, and DeepSeek's DeepSeek-V3 and DeepSeek-R1 as the corresponding System 1 and System 2 pairs, respectively. Inspired by recent research~\cite{wang2024tabletime,hoo2025tabular,hu2025contextalignment}, we reformulate numerical time series into a tabular format, i.e., "timestamp : numerical value", to enable LLMs as powerful time-series analysts.
\subsection{Evaluation Module}
We comprehensively consider the following four common settings: unimodal short-term, unimodal long-term, multi-modal short-term, and multi-modal long-term. In multi-modal TSF, both numerical series and aligned textual context series are used as inputs, whereas unimodal TSF uses only numerical series. The forecasting period for long-term TSF is the next six months whereas the short-term TSF is the next three months. We follow most existing TSF works~\cite{wu2021autoformer,wutimesnet,nietime} by setting the lookback window length to 96 by default. 
We use Mean Squared Error (MSE) as the evaluation metric. To avoid data contamination—meaning the evaluation dataset may have been seen during the foundation model's pretraining—we use horizon windows after October 2023, which is the knowledge cutoff date of selected foundation models.

%% file: Section/New_Exp_1.tex
\begin{table*}[t] 
    \centering
    \caption{Results with OpenAI's System 1 (GPT-4o) and 2 (o1-mini) Models. We report the mean MSE and standard deviation over three repeated experiments. Reasoning strategies that outperform the direct System 1 are highlighted in \better{green}, while those that perform worse or have similar performance (due to higher computational cost) are marked in \worse{red}. In "Win System 1," we present the probability of each reasoning strategy outperforming System 1 across datasets. We observe that \better{only the self-consistency strategy is consistently effective}, while \worse{the System 2 strategy is consistently ineffective}.}
    \vspace{-3mm}
\begin{subtable}{\textwidth}
\centering
\caption{Results of Unimodal Short-term TSF Settings. We use numerical series only to forecast the next three months.}
\begin{tabular}{c|c|ccc|c} 
\hline
\multirow{2}{*}{Dataset}           & System 1    & \multicolumn{3}{c|}{~System 1 with~Test-time Reasoning Enhancement} & System~~~~ 2  \\ 
\cline{2-6}
                                   & GPT-4o      & with CoT     & with Self-Consistency & with Self-Correction  & o1-mini       \\ 
\hline
Agriculture & 0.021±0.011 & \worse{0.909±1.275} & \better{0.021±0.003} & \worse{0.025±0.007} & \worse{0.069±0.013} \\
 Climate & 1.599±0.500 & \worse{1.704±0.164} & \better{1.517±0.263} & \worse{1.998±0.677} & \better{1.412±0.159} \\
 Economy & 0.631±0.135 & \worse{0.638±0.410} & \better{0.450±0.171} & \worse{1.018±0.184} & \better{0.583±0.001} \\
 Energy & 0.363±0.110 & \better{0.258±0.029} & \better{0.167±0.242} & \worse{0.396±0.086} & \worse{0.930±0.747} \\
 Flu & 0.568±0.425 & \worse{0.592±0.291} & \better{0.481±0.288} & \worse{0.663±0.078} & \worse{1.441±1.234} \\
 Security & 0.093±0.029 & \worse{0.259±0.001} & \better{0.084±0.028} & \worse{0.165±0.070} & \worse{0.225±0.048} \\
 Employment & 0.010±0.004 & \better{0.006±0.002} & \worse{0.012±0.001} & \worse{0.013±0.003} & \worse{0.021±0.003} \\
Traffic & 0.385±0.471 & \better{0.113±0.063} & \worse{0.047±0.009} & \better{0.053±0.009} & \worse{0.566±0.731} \\
\hline
Win System 1 & NA & \worse{3/8} & \better{5/8} & \worse{1/8} & \worse{2/8} \\
\hline
\end{tabular}
\end{subtable}
\vspace{2mm}
\begin{subtable}{\textwidth}
\centering
\caption{Results of Multimodal Short-term TSF Settings. We use numerical series with textual context series to forecast the next three months.}
\begin{tabular}{c|c|ccc|c} 
\hline
\multirow{2}{*}{Dataset}           & System 1    & \multicolumn{3}{c|}{~System 1 with~Test-time Reasoning Enhancement} & System~~~~ 2  \\ 
\cline{2-6}
                                   & GPT-4o      & with CoT     & with Self-Consistency & with Self-Correction  & o1-mini       \\ 
\hline
 Agriculture & 0.018±0.015 & \better{0.018±0.011} & \better{0.013±0.008} & \better{0.018±0.006} & \worse{0.045±0.056} \\
 Climate & 1.716±0.580 & \worse{1.920±0.505} & \better{1.712±0.191} & \worse{2.042±0.609} & \better{1.603±0.496} \\
 Economy & 0.569±0.162 & \worse{0.940±0.445} & \better{0.291±0.127} & \better{0.503±0.071} & \worse{0.583±0.001} \\
 Energy & 0.541±0.457 & \better{0.316±0.125} & \better{0.187±0.090} & \better{0.225±0.080} & \worse{0.718±0.786} \\
 Flu & 0.548±0.164 & \worse{1.071±0.643} & \better{0.288±0.071} & \worse{1.261±1.164} & \worse{0.983±1.177} \\
 Security & 0.076±0.052 & \worse{0.110±0.087} & \worse{0.146±0.025} & \worse{0.151±0.035} & \worse{0.244±0.020} \\
 Employment & 0.020±0.006 & \worse{0.020±0.003} & \better{0.019±0.003} & \worse{0.021±0.004} & \worse{0.028±0.008} \\
Traffic & 0.551±0.396 & \worse{1.577±1.421} & \better{0.030±0.010} & \better{0.347±0.349} & \worse{0.911±0.594} \\
\hline
Win System 1 & NA & \worse{2/8} & \better{7/8} & \worse{4/8} & \worse{1/8} \\
\hline
\end{tabular}
\end{subtable}
\vspace{2mm}
\begin{subtable}{\textwidth}
\centering
\caption{Results of Unimodal Long-term TSF Settings. We use numerical series only to forecast the next six months.}
\begin{tabular}{c|c|ccc|c} 
\hline
\multirow{2}{*}{Dataset}           & System 1    & \multicolumn{3}{c|}{~System 1 with~Test-time Reasoning Enhancement} & System~~~~ 2  \\ 
\cline{2-6}
                                   & GPT-4o      & with CoT     & with Self-Consistency & with Self-Correction  & o1-mini       \\ 
\hline
 Agriculture & 0.093±0.057 & \worse{0.920±1.134} & \better{0.057±0.011} & \better{0.068±0.018} & \worse{0.293±0.089} \\
 Climate & 0.754±0.051 & \worse{1.199±0.132} & \worse{0.811±0.081} & \worse{0.877±0.041} & \better{0.708±0.058} \\
 Economy & 0.463$\pm$0.146 & \worse{1.040$\pm$0.482} & \worse{0.620$\pm$0.116} & \worse{0.748$\pm$0.069} & \better{0.359$\pm$0.001} \\
 Energy & 0.197$\pm$0.038 & \worse{0.746$\pm$0.500} & \better{0.177$\pm$0.062} & \worse{0.296$\pm$0.153} & \worse{0.926$\pm$0.771} \\
 Flu & 0.219$\pm$0.053 & \worse{0.967$\pm$0.412} & \worse{0.230$\pm$0.077} & \worse{0.639$\pm$0.479} & \worse{0.862$\pm$0.597} \\
 Security & 0.183$\pm$0.044 & \better{0.162$\pm$0.038} & \better{0.135$\pm$0.011} & \better{0.165$\pm$0.017} & \worse{0.211$\pm$0.075} \\
 Employment & 0.011$\pm$0.006 & \worse{0.013$\pm$0.002} & \better{0.009$\pm$0.003} & \worse{0.013$\pm$0.004} & \worse{0.053$\pm$0.015} \\
Traffic & 0.066$\pm$0.046 & \worse{0.218$\pm$0.158} & \better{0.046$\pm$0.016} & \better{0.036$\pm$0.008} & \worse{0.091$\pm$0.042} \\
\hline
Win System 1 & NA &  \worse{1/8} &  \better{5/8} &  \worse{3/8} &  \worse{2/8} \\
\hline
\end{tabular}
\end{subtable}
\vspace{2mm}
\begin{subtable}{\textwidth}
\centering
\caption{Results of Multimodal Long-term TSF Settings. We use numerical series with textual context series to forecast the next six months.}
\begin{tabular}{c|c|ccc|c} 
\hline
\multirow{2}{*}{Dataset}           & System 1    & \multicolumn{3}{c|}{~System 1 with~Test-time Reasoning Enhancement} & System~~~~ 2  \\ 
\cline{2-6}
                                   & GPT-4o      & with CoT     & with Self-Consistency & with Self-Correction  & o1-mini       \\ 
\hline
 Agriculture & 0.110±0.065 & \better{0.097±0.044} & \better{0.063±0.009} & \better{0.051±0.042} & \worse{0.210±0.022} \\
 Climate & 1.365±0.479 & \better{0.995±0.109} & \better{1.065±0.014} & \better{0.912±0.004} & \worse{1.549±0.566} \\
 Economy & 0.487±0.237 & \worse{1.027±0.321} & \worse{0.500±0.184} & \worse{0.543±0.074} & \worse{0.827±0.662} \\
 Energy & 0.365±0.185 & \better{0.254±0.122} & \worse{33.743±23.911} & \better{0.293±0.026} & \worse{0.707±0.499} \\
 Flu & 0.291±0.065 & \worse{0.369±0.058} & \worse{0.445±0.210} & \worse{0.529±0.365} & \worse{1.070±0.284} \\
 Security & 0.196±0.056 & \better{0.188±0.027} & \better{0.140±0.028} & \better{0.116±0.041} & \worse{0.207±0.001} \\
 Employment & 0.015±0.002 & \worse{0.021±0.007} & \worse{0.021±0.002} & \worse{0.106±0.115} & \worse{0.031±0.003} \\
Traffic & 0.207±0.205 & \worse{0.341±0.402} & \better{0.045±0.013} & \worse{0.377±0.504} & \worse{1.482±1.788} \\
\hline
Win System 1 & NA & \worse{4/8} & \worse{4/8} & \worse{4/8} & \worse{0/8} \\
\hline
\end{tabular}
\end{subtable}
     \label{tab:GPT-A}
\end{table*}

%% file: Section/New_Exp_2.tex
\begin{table*}[t] 
    \centering
    \caption{Results with Google's System 1 (Gemini-2.0-flash) and 2 (Gemini-2.0-flash-thinking) Models. We report the mean MSE and standard deviation over three repeated experiments. Reasoning strategies that outperform the direct System 1 are highlighted in \better{green}, while those that perform worse or have similar performance (due to higher computational cost) are marked in \worse{red}. In "Win System 1," we present the probability of each reasoning strategy outperforming System 1 across datasets. We again observe that \better{only the self-consistency strategy consistently works}, while \worse{the System 2 strategy consistently fails}.}
    \vspace{-3mm}
\begin{subtable}{\textwidth}
\centering
\caption{Results of Unimodal Short-term TSF Settings. We use numerical series only to forecast the next three months.}
\begin{tabular}{c|c|ccc|c} 
\hline
\multirow{2}{*}{Dataset}           & System 1    & \multicolumn{3}{c|}{~System 1 with~Test-time Reasoning Enhancement} & System~~~~ 2  \\ 
\cline{2-6}
                                   & Gemini-2.0-flash      & with CoT     & with Self-Consistency & with Self-Correction  & Gemini-2.0-flash-thinking       \\ 
\hline
 Agriculture & 0.011$\pm$0.001 & \better{0.010$\pm$0.004} & \better{0.009$\pm$0.004} & \worse{0.012$\pm$0.008} & \worse{0.017$\pm$0.004} \\
 Climate & 1.234$\pm$0.239 & \worse{1.800$\pm$0.326} & \worse{1.749$\pm$0.791} & \worse{1.703$\pm$0.280} & \worse{2.416$\pm$0.112} \\
 Economy & 0.113$\pm$0.007 & \worse{0.272$\pm$0.256} & \worse{0.229$\pm$0.145} & \worse{0.121$\pm$0.026} & \worse{0.172$\pm$0.049} \\
 Energy & 0.172$\pm$0.038 & \worse{0.181$\pm$0.048} & \better{0.132$\pm$0.047} & \worse{0.235$\pm$0.060} & \worse{0.327$\pm$0.054} \\
 Flu & 0.809$\pm$0.353 & \better{0.641$\pm$0.224} & \better{0.402$\pm$0.197} & \worse{1.854$\pm$1.271} & \worse{2.068$\pm$1.076} \\
 Security & 0.170$\pm$0.054 & \worse{0.252$\pm$0.104} & \worse{0.380$\pm$0.323} & \worse{0.191$\pm$0.095} & \worse{0.259$\pm$0.001} \\
 Employment & 0.002$\pm$0.001 & \worse{0.005$\pm$0.003} & \worse{0.004$\pm$0.004} & \worse{0.004$\pm$0.002} & \worse{0.311$\pm$0.001} \\
Traffic & 0.347$\pm$0.415 & \better{0.097$\pm$0.060} & \better{0.016$\pm$0.006} & \better{0.034$\pm$0.014} & \better{0.201$\pm$0.001} \\
\hline
Win System 1 & NA & $\worse{3/8}$ & $\worse{4/8}$ & $\worse{1/8}$ & $\worse{1/8}$ \\
\hline
\end{tabular}
\end{subtable}
\vspace{1mm}
\begin{subtable}{\textwidth}
\centering
\caption{Results of Multimodal Short-term TSF Settings. We use numerical series with textual context series to forecast the next three months.}
\begin{tabular}{c|c|ccc|c} 
\hline
\multirow{2}{*}{Dataset}           & System 1    & \multicolumn{3}{c|}{~System 1 with~Test-time Reasoning Enhancement} & System~~~~ 2  \\ 
\cline{2-6}
                                   & Gemini-2.0-flash      & with CoT     & with Self-Consistency & with Self-Correction  & Gemini-2.0-flash-thinking       \\ 
\hline

 Agriculture & 0.010$\pm$0.003 & \better{0.006$\pm$0.001} & \better{0.009$\pm$0.002} & \worse{0.011$\pm$0.004} & \better{0.008$\pm$0.002} \\
 Climate & 2.115$\pm$0.660 & \better{1.725$\pm$0.227} & \better{1.980$\pm$0.760} & \better{1.529$\pm$0.290} & \better{2.106$\pm$0.294} \\
 Economy & 0.376$\pm$0.085 & \better{0.326$\pm$0.067} & \better{0.373$\pm$0.079} & \better{0.283$\pm$0.083} & \worse{0.509$\pm$0.109} \\
 Energy & 0.143$\pm$0.069 & \better{0.117$\pm$0.015} & \worse{0.143$\pm$0.027} & \better{0.091$\pm$0.065} & \worse{0.218$\pm$0.106} \\
 Flu & 0.594$\pm$0.219 & \worse{0.607$\pm$0.294} & \better{0.332$\pm$0.102} & \worse{1.422$\pm$0.542} & \worse{3.171$\pm$0.001} \\
 Security & 0.558$\pm$0.604 & \better{0.145$\pm$0.050} & \better{0.172$\pm$0.119} & \better{0.141$\pm$0.065} & \better{0.259$\pm$0.001} \\
 Employment & 0.013$\pm$0.002 & \worse{0.015$\pm$0.002} & \better{0.011$\pm$0.002} & \better{0.011$\pm$0.003} & \worse{0.311$\pm$0.001} \\
Traffic & 0.322$\pm$0.196 & \better{0.046$\pm$0.017} & \better{0.163$\pm$0.106} & \worse{0.425$\pm$0.235} & \better{0.201$\pm$0.001} \\
\hline
Win System 1 & NA & $\better{6/8}$ & $\better{7/8}$ & $\better{5/8}$ & $\worse{4/8}$ \\
\hline
\end{tabular}
\end{subtable}
\vspace{1mm}
\begin{subtable}{\textwidth}
\centering
\caption{Results of Unimodal Long-term TSF Settings. We use numerical series only to forecast the next six months.}
\begin{tabular}{c|c|ccc|c} 
\hline
\multirow{2}{*}{Dataset}           & System 1    & \multicolumn{3}{c|}{~System 1 with~Test-time Reasoning Enhancement} & System~~~~ 2  \\ 
\cline{2-6}
                                   & Gemini-2.0-flash      & with CoT     & with Self-Consistency & with Self-Correction  & Gemini-2.0-flash-thinking       \\ 
\hline
 Agriculture & 0.032$\pm$0.007 & \worse{0.036$\pm$0.011} & \worse{0.035$\pm$0.007} & \worse{0.077$\pm$0.026} & \worse{0.093$\pm$0.018} \\
 Climate & 1.476$\pm$0.651 & \better{0.964$\pm$0.321} & \better{0.674$\pm$0.092} & \better{0.908$\pm$0.153} & \better{1.240$\pm$0.705} \\
 Economy & 0.092$\pm$0.038 & \worse{0.216$\pm$0.142} & \better{0.078$\pm$0.013} & \better{0.066$\pm$0.003} & \worse{0.244$\pm$0.035} \\
 Energy & 0.303$\pm$0.044 & \better{0.130$\pm$0.021} & \better{0.241$\pm$0.060} & \worse{0.489$\pm$0.134} & \better{0.241$\pm$0.148} \\
 Flu & 1.190$\pm$1.171 & \better{1.049$\pm$0.447} & \better{0.596$\pm$0.128} & \better{1.095$\pm$0.546} & \worse{1.920$\pm$0.001} \\
 Security & 0.196$\pm$0.052 & \worse{0.533$\pm$0.493} & \worse{0.955$\pm$0.389} & \better{0.154$\pm$0.031} & \worse{0.207$\pm$0.001} \\
 Employment & 0.011$\pm$0.001 & \worse{0.019$\pm$0.007} & \better{0.009$\pm$0.002} & \worse{0.013$\pm$0.005} & \worse{0.268$\pm$0.001} \\
Traffic & 0.068$\pm$0.063 & \worse{0.215$\pm$0.079} & \worse{0.074$\pm$0.048} & \better{0.050$\pm$0.013} & \worse{0.414$\pm$0.001} \\
\hline
Win System 1& NA & $\worse{3/8}$ & $\better{5/8}$ & $\better{5/8}$ & $\worse{2/8}$ \\
\hline
\end{tabular}
\end{subtable}
\vspace{1mm}
\begin{subtable}{\textwidth}
\centering
\caption{Results of Multimodal Long-term TSF Settings. We use numerical series with textual context series to forecast the next six months.}
\begin{tabular}{c|c|ccc|c} 
\hline
\multirow{2}{*}{Dataset}           & System 1    & \multicolumn{3}{c|}{~System 1 with~Test-time Reasoning Enhancement} & System~~~~ 2  \\ 
\cline{2-6}
                                   & Gemini-2.0-flash      & with CoT     & with Self-Consistency & with Self-Correction  & Gemini-2.0-flash-thinking       \\ 
\hline
 Agriculture & 0.052$\pm$0.026 & \better{0.034$\pm$0.009} & \better{0.034$\pm$0.006} & \better{0.024$\pm$0.007} & \worse{0.096$\pm$0.032} \\
 Climate & 1.644$\pm$0.398 & \better{1.452$\pm$0.461} & \better{1.318$\pm$0.079} & \better{1.292$\pm$0.401} & \better{1.006$\pm$0.327} \\
 Economy & 0.092$\pm$0.010 & \worse{0.234$\pm$0.049} & \worse{0.134$\pm$0.044} & \worse{10.357$\pm$14.475} & \worse{1.093$\pm$0.806} \\
 Energy & 0.138$\pm$0.106 & \worse{0.208$\pm$0.116} & \worse{0.159$\pm$0.077} & \worse{0.384$\pm$0.074} & \worse{0.713$\pm$0.513} \\
 Flu & 0.659$\pm$0.173 & \better{0.557$\pm$0.164} & \better{0.477$\pm$0.006} & \worse{0.785$\pm$0.064} & \worse{1.920$\pm$0.001} \\
 Security & 0.123$\pm$0.062 & \better{0.109$\pm$0.020} & \worse{0.142$\pm$0.043} & \worse{0.151$\pm$0.053} & \worse{0.207$\pm$0.001} \\
 Employment & 0.029$\pm$0.004 & \better{0.022$\pm$0.003} & \better{0.026$\pm$0.003} & \better{0.026$\pm$0.002} & \worse{0.268$\pm$0.001} \\
Traffic & 0.085$\pm$0.068 & \better{0.037$\pm$0.027} & \better{0.020$\pm$0.007} & \better{0.058$\pm$0.010} & \worse{0.414$\pm$0.001} \\
\hline
Win System 1 & NA & $\better{6/8}$ & $\better{5/8}$ & $\worse{4/8}$ & $\worse{1/8}$ \\
\hline
\end{tabular}
\end{subtable}
     \label{tab:Gemini-2}
\end{table*}

%% file: Section/New_Exp_3.tex
\begin{table*}[t] 
    \centering
    \caption{Results with DeepSeek's System 1 (DeepSeek-V3) and 2 (DeepSeek-R1) Models. We report the mean MSE and standard deviation over three repeated experiments. Reasoning strategies that outperform the direct System 1 are highlighted in \better{green}, while those that perform worse or have similar performance (due to higher computational cost) are marked in \worse{red}. In "Win System 1," we present the probability of each reasoning strategy outperforming System 1 across datasets. We observe that \better{the self-consistency still consistently works}. We find that \better{\uline{DeepSeek-R1 is the only System 2 model that is effective}} for TSF, which we attribute to its Group Relative Policy Optimization approach aligning well with the TSF task.}
    \vspace{-3mm}
\begin{subtable}{\textwidth}
\centering
\caption{Results of Unimodal Short-term TSF Settings. We use numerical series only to forecast the next three months.}
\begin{tabular}{c|c|ccc|c} 
\hline
\multirow{2}{*}{Dataset}           & System 1    & \multicolumn{3}{c|}{~System 1 with~Test-time Reasoning Enhancement} & System~~~~ 2  \\ 
\cline{2-6}
                                   & DeepSeek-V3      & with CoT     & with Self-Consistency & with Self-Correction  & DeepSeek-R1       \\ 
\hline
 Agriculture & 0.038$\pm$0.032 & \better{0.019$\pm$0.001} & \worse{0.046$\pm$0.015} & \better{0.013$\pm$0.003} & \better{0.016$\pm$0.010} \\
 Climate & 1.216$\pm$0.202 & \worse{2.650$\pm$0.905} & \better{1.207$\pm$0.197} & \worse{1.246$\pm$0.081} & \worse{1.541$\pm$0.397} \\
 Economy & 0.406$\pm$0.218 & \worse{0.433$\pm$0.031} & \better{0.284$\pm$0.227} & \worse{0.441$\pm$0.161} & \worse{0.583$\pm$0.001} \\
 Energy & 0.736$\pm$0.752 & \better{0.212$\pm$0.022} & \better{0.187$\pm$0.011} & \better{0.182$\pm$0.063} & \better{0.189$\pm$0.021} \\
 Flu & 1.464$\pm$1.031 & \worse{1.650$\pm$0.236} & \better{0.980$\pm$0.445} & \worse{1.682$\pm$0.292} & \better{1.298$\pm$1.330} \\
 Security & 0.283$\pm$0.140 & \better{0.218$\pm$0.093} & \better{0.185$\pm$0.052} & \better{0.116$\pm$0.012} & \better{0.247$\pm$0.017} \\
 Employment & 0.036$\pm$0.019 & \better{0.020$\pm$0.006} & \better{0.035$\pm$0.019} & \better{0.018$\pm$0.007} & \better{0.012$\pm$0.005} \\
Traffic & 0.066±0.031 & \worse{0.201±0.001} & \worse{0.109±0.028} & \worse{0.107±0.067} & \worse{0.113±0.073} \\
\hline
Win System 1 & NA & \worse{4/8} & \better{6/8} & \worse{4/8} & \better{5/8}\\
\hline
\end{tabular}
\end{subtable}

\begin{subtable}{\textwidth}
\centering
\caption{Results of Multimodal Short-term TSF Settings. We use numerical series with textual context series to forecast the next three months.}
\begin{tabular}{c|c|ccc|c} 
\hline
\multirow{2}{*}{Dataset}           & System 1    & \multicolumn{3}{c|}{~System 1 with~Test-time Reasoning Enhancement} & System~~~~ 2  \\ 
\cline{2-6}
                                   & DeepSeek-V3      & with CoT     & with Self-Consistency & with Self-Correction  & DeepSeek-R1       \\ 
\hline
 Agriculture & 0.032$\pm$0.012 & \better{0.027$\pm$0.006} & \better{0.023$\pm$0.001} & \worse{0.042$\pm$0.025} & \worse{2.712$\pm$0.001} \\
 Climate & 1.428$\pm$0.432 & \worse{1.857$\pm$0.431} & \better{1.371$\pm$0.001} & \better{1.411$\pm$0.258} & \worse{2.235$\pm$0.850} \\
 Economy & 0.427$\pm$0.174 & \worse{0.598$\pm$0.069} & \better{0.306$\pm$0.005} & \better{0.369$\pm$0.128} & \worse{0.615$\pm$0.101} \\
 Energy & 0.253$\pm$0.089 & \worse{0.486$\pm$0.318} & \better{0.197$\pm$0.001} & \worse{0.505$\pm$0.339} & \worse{0.731$\pm$0.777} \\
 Flu & 1.073$\pm$0.447 & \worse{1.564$\pm$0.982} & \better{0.362$\pm$0.161} & \better{0.441$\pm$0.173} & \worse{1.329$\pm$1.306} \\
 Security & 0.186$\pm$0.001 & \worse{0.206$\pm$0.010} & \worse{0.187$\pm$0.001} & \better{0.130$\pm$0.018} & \better{0.161$\pm$0.051} \\
 Employment & 0.016$\pm$0.001 & \worse{0.022$\pm$0.003} & \worse{0.016$\pm$0.001} & \worse{0.016$\pm$0.001} & \worse{0.114$\pm$0.139} \\
Traffic & 0.201$\pm$0.001 & \worse{0.201$\pm$0.001} & \worse{0.201$\pm$0.001} & \better{0.114$\pm$0.063} & \better{0.153$\pm$0.069} \\
\hline
Win System 1 & NA & $\worse{1/8}$ & $\better{5/8}$ & $\better{5/8}$ & $\worse{2/8}$ \\
\hline
\end{tabular}
\end{subtable}

\begin{subtable}{\textwidth}
\centering
\caption{Results of Unimodal Long-term TSF Settings. We use numerical series only to forecast the next six months.}
\begin{tabular}{c|c|ccc|c} 
\hline
\multirow{2}{*}{Dataset}           & System 1    & \multicolumn{3}{c|}{~System 1 with~Test-time Reasoning Enhancement} & System~~~~ 2  \\ 
\cline{2-6}
                                   & DeepSeek-V3      & with CoT     & with Self-Consistency & with Self-Correction  & DeepSeek-R1       \\ 
\hline
 Agriculture & 0.216$\pm$0.049 & \better{0.102$\pm$0.034} & \better{0.103$\pm$0.014} & \better{0.121$\pm$0.065} & \better{0.091$\pm$0.019} \\
 Climate & 0.902$\pm$0.001 & \worse{1.383$\pm$0.227} & \better{0.786$\pm$0.153} & \worse{0.913$\pm$0.078} & \better{0.662$\pm$0.051} \\
 Economy & 0.613$\pm$0.776 & \better{0.540$\pm$0.386} & \better{0.393$\pm$0.113} & \worse{0.948$\pm$0.589} & \better{0.359$\pm$0.001} \\
 Energy & 0.603$\pm$0.359 & \better{0.575$\pm$0.452} & \worse{0.923$\pm$0.265} & \better{0.332$\pm$0.150} & \worse{1.396$\pm$0.001} \\
 Flu & 0.841$\pm$0.215 & \better{0.658$\pm$0.227} & \better{0.538$\pm$0.021} & \worse{0.939$\pm$0.328} & \worse{0.972$\pm$0.533} \\
 Security & 0.275$\pm$0.060 & \better{0.245$\pm$0.039} & \worse{0.280$\pm$0.004} & \better{0.186$\pm$0.033} & \better{0.168$\pm$0.028} \\
 Employment & 0.051$\pm$0.013 & \better{0.021$\pm$0.002} & \better{0.039$\pm$0.006} & \better{0.023$\pm$0.003} & \better{0.021$\pm$0.001} \\
Traffic & 0.414$\pm$0.001 & \better{0.209$\pm$0.145} & \worse{94.305$\pm$66.620} & \better{0.306$\pm$0.153} & \better{0.158$\pm$0.181} \\
\hline
Win System 1 & NA & \better{7/8} & \better{5/8} & \better{5/8} & \better{6/8} \\
\hline
\end{tabular}
\end{subtable}

\begin{subtable}{\textwidth}
\centering
\caption{Results of Multimodal Long-term TSF Settings. We use numerical series with textual context series to forecast the next six months.}
\begin{tabular}{c|c|ccc|c} 
\hline
\multirow{2}{*}{Dataset}           & System 1    & \multicolumn{3}{c|}{~System 1 with~Test-time Reasoning Enhancement} & System~~~~ 2  \\ 
\cline{2-6}
                                   & DeepSeek-V3      & with CoT     & with Self-Consistency & with Self-Correction  & DeepSeek-R1       \\ 
\hline
 Agriculture & 0.088$\pm$0.058 & \better{0.063$\pm$0.022} & \worse{0.136$\pm$0.080} & \worse{0.119$\pm$0.078} & \better{0.019$\pm$0.010} \\
 Climate & 0.897$\pm$0.001 & \worse{2.193$\pm$0.330} & \worse{0.897$\pm$0.001} & \worse{0.939$\pm$0.074} & \worse{1.849$\pm$0.570} \\
 Economy & 0.629$\pm$0.147 & \better{0.558$\pm$0.282} & \better{0.486$\pm$0.074} & \better{0.623$\pm$0.218} & \worse{0.806$\pm$0.354} \\
 Energy & 0.995$\pm$0.139 & \worse{1.286$\pm$0.568} & \better{0.809$\pm$0.241} & \better{0.493$\pm$0.112} & \better{0.746$\pm$0.459} \\
 Flu & 2.624$\pm$2.400 & \better{0.974$\pm$0.446} & \better{0.644$\pm$0.488} & \better{1.135$\pm$0.643} & \better{1.560$\pm$0.957} \\
 Security & 0.179$\pm$0.002 & \worse{0.250$\pm$0.024} & \better{0.156$\pm$0.027} & \worse{0.274$\pm$0.071} & \better{0.134$\pm$0.055} \\
 Employment & 0.034$\pm$0.001 & \better{0.029$\pm$0.008} & \worse{0.034$\pm$0.001} & \better{0.030$\pm$0.005} & \worse{0.105$\pm$0.115} \\
Traffic & 0.414$\pm$0.001 & \worse{0.414$\pm$0.001} & \worse{0.414$\pm$0.001} & \better{0.192$\pm$0.157} & \better{0.152$\pm$0.185} \\
\hline
Win System 1 & NA & $\worse{4/8}$ & $\worse{4/8}$ & $\better{5/8}$ & $\better{5/8}$ \\
\hline
\end{tabular}
\end{subtable}
\label{tab:deepseek-3}
\end{table*}

\clearpage
\begin{figure}[t]
    \centering
    \includegraphics[width=0.5\textwidth]{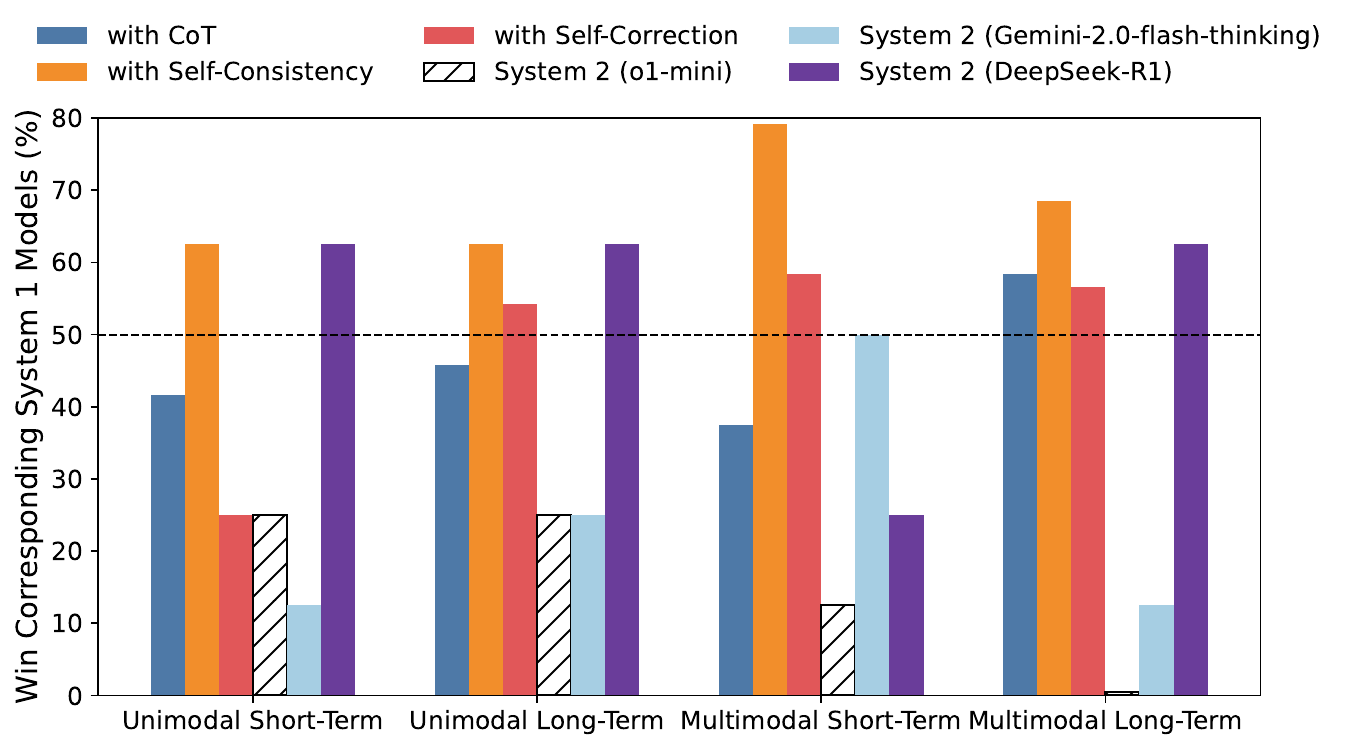}  % 或 tsf_comparison.pdf
    \vspace{-3mm}
    \caption{The average win rate of reasoning strategies compared to corresponding direct System 1 across all datasets and settings. We observe the consistent and significant effectiveness of self-consistency, as well as the unique effectiveness of DeepSeek-R1 among System 2 strategies.}
    \label{fig:tsf_comparison-A}
        \vspace{-5mm}
\end{figure}
\section{Experimental Results and Insights}
Based on the constructed \method suite, we conduct experiments to evaluate reasoning strategies for zero-shot TSF across eight datasets and four settings. We repeat each experiment three times, reporting the average MSE and standard deviation. We detail the experimental results in Table~\ref{tab:GPT-A}, ~\ref{tab:Gemini-2}, and ~\ref{tab:deepseek-3} corresponding to OpenAI's, Google's, and DeepSeek's foundation models. We visualize the average win rate of different reasoning strategies relative to the direct System 1, where 50\% means a tie, in Figure~\ref{fig:tsf_comparison-A}. We then discuss the two research questions raised in Section~\ref{sec:intro} one by one, following the structure of Answer – Evidence – Analysis. 
\subsection{RQ1: Can TSF Benefit from Reasoning?}
\observationbox{Overall Answer: TSF can benefit from enhanced reasoning ability}{We observe that in all four TSF scenarios, at least two reasoning strategies are effective, by outperforming the corresponding System 1 models in over 50\% of case; at least one reasoning strategy is significant, by surpassing the corresponding System 1 model in over 60\% cases.} 

\observationbox{From Short-term vs. Long-term Perspective: Long-term TSF benefits more consistently.}{We observe that long-term TSF, in both unimodal and multimodal settings, consistently benefits from all three System 1-based reasoning strategies across datasets and methods. Specifically, the CoT, Self-Consistency, and Self-Correction strategies outperform System 1 models in 52.08\%, 58.33\%, and 54.17\% cases, respectively. In contrast, short-term TSF only consistently benefits from the Self-Consistency strategy. This aligns with TSF, where long-term forecasting requires more consideration of temporal and event influences, while short-term forecasting is more similar to the lookback window.} 

\observationbox{From Unimodal vs. Multimodal Perspective: Multimodal TSF benefits more significantly.}{We observe that multimodal TSF, in both long-term and short-term settings, benefits more significantly from reasoning enhancement. Specifically, the Self-Consistency and Self-Correction strategies outperform System 1 models in  66.67\% and 56.25\% cases, respectively. In contrast, unimodal TSF only significantly benefits from the Self-Consistency strategy. This aligns with the intuition that multimodal TSF, which provides textual context for forecasting, requires more reasoning.} 

\subsection{RQ2: What Reasoning Strategies TSF Need?}
\observationbox{Overall Answer: self-consistency is the current best.}{We observe that the self-consistency strategy is effective and outperforms the System 1 model at a rate of 60\% to 80\%. We believe that self-consistency works by selecting the most coherent reasoning path from various options, which follows the inherent logic of TSF: to consider multiple possible future scenarios and choose the most likely one for prediction.}

\rejectionbox{From System 1 vs System 2 Perspective: Reasoning Enhanced System 1 win}{We observe that System 1 with test-time reasoning enhancement achieves an average effectiveness of 66.67\%, which is much higher than the 33.33\% of System 2. This suggests that pure System 2 reasoning may not be the correct answer for TSF. In contrast, reasoning-enhanced System 1 is more suitable, as it combines quick responses with slow thinking in line with TSF, which also combines superfacial pattern recognition, especially periodicity and trends~\cite{cleveland1990stl,liu2024lstprompt}, and deep reasoning, especially event influence~\cite{liu2025time}}

\observationbox{From System 2 Perspective: DeepSeek-R1 is the only effective reasoning model.}{We observe that DeepSeek-R1 is the only effective model, while the other two, o1-mini and Gemini-2.0-Flash-Thinking, are ineffective. DeepSeek-R1 shows significant improvements in three out of four settings, surpassing the System 1 model (DeepSeek-V3) in 60\% cases. We believe this is due to DeepSeek-R1's unique reinforcement learning approach, called Group Relative Policy Optimization (GRPO)~\cite{guo2025deepseek}, which focuses solely on outcomes rather than on labeled reasoning paths. Clearly, for TSF, relying on precise reasoning paths to forecast uncertain future numerical series is also not rational.}

%% file: Section/Toolkits.tex
\section{Open-Source Toolkits: Evaluation Suite, Test-Time Scaling Law, and Datasets}
We provide three key open-source toolkits to support future research on foundational TSF reasoning models. Specifically, these include our \method as an easy-to-use evaluation suite with experiment logs, a newly verified test-time scaling law on foundation time-series models for zero-shot TSF, and the first TSF dataset with reasoning annotations distilled from six advanced foundation models, named \data. All resources are publicly accessible at: \url{https://github.com/AdityaLab/OpenTimeR}
% \subsection{Released Evaluation Suite}
\vspace{0.1in}
\par\noindent\textbf{Released Evaluation Suite.}
We fully release \method as an easy-to-use evaluation suite to facilitate future research, including the code, documents, and experiment logs. The released code supports batch experiments, unified selection of reasoning strategy with LLMs, and automated saving and extraction of experimental results. We provide the hyperparameters and model outputs of about 1500 experiments in our benchmarking.
\vspace{0.05in}

% \subsection{Test-Time Scaling Law Verification}
\begin{figure}[t]
    \centering
    % \vspace{-0.05in}
    \includegraphics[width=0.5\textwidth]{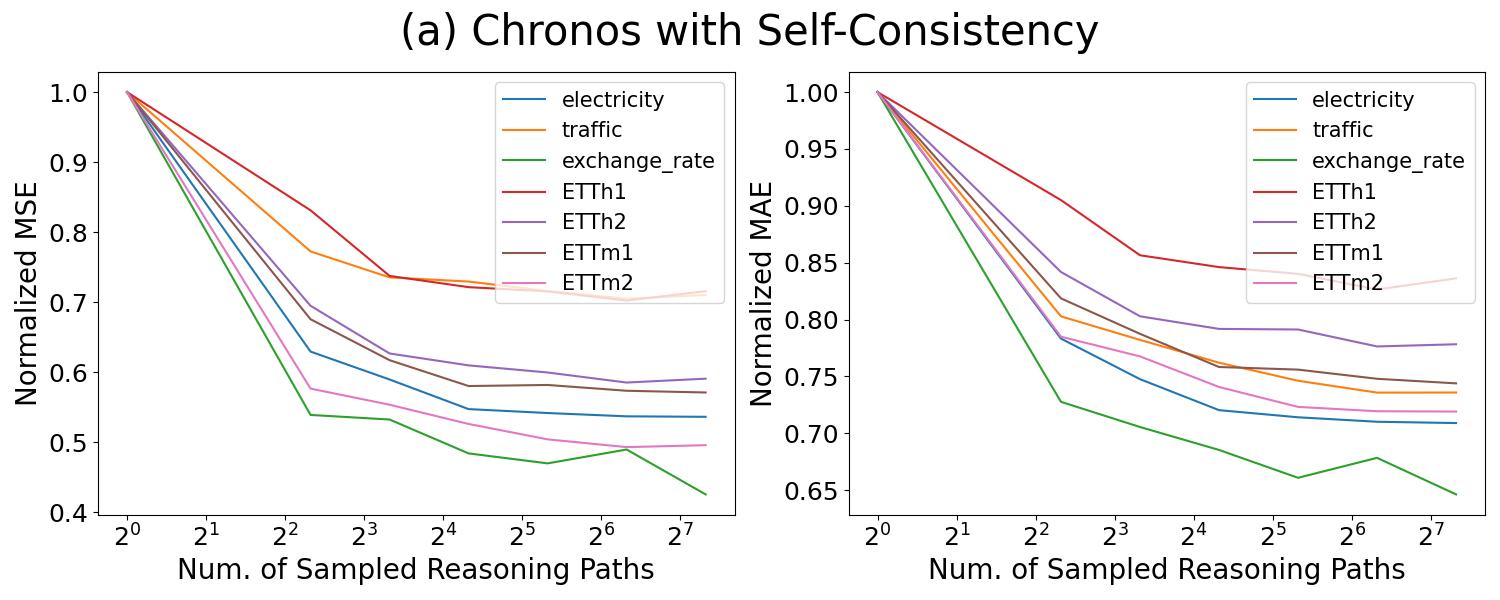} \\
    \includegraphics[width=0.5\textwidth]{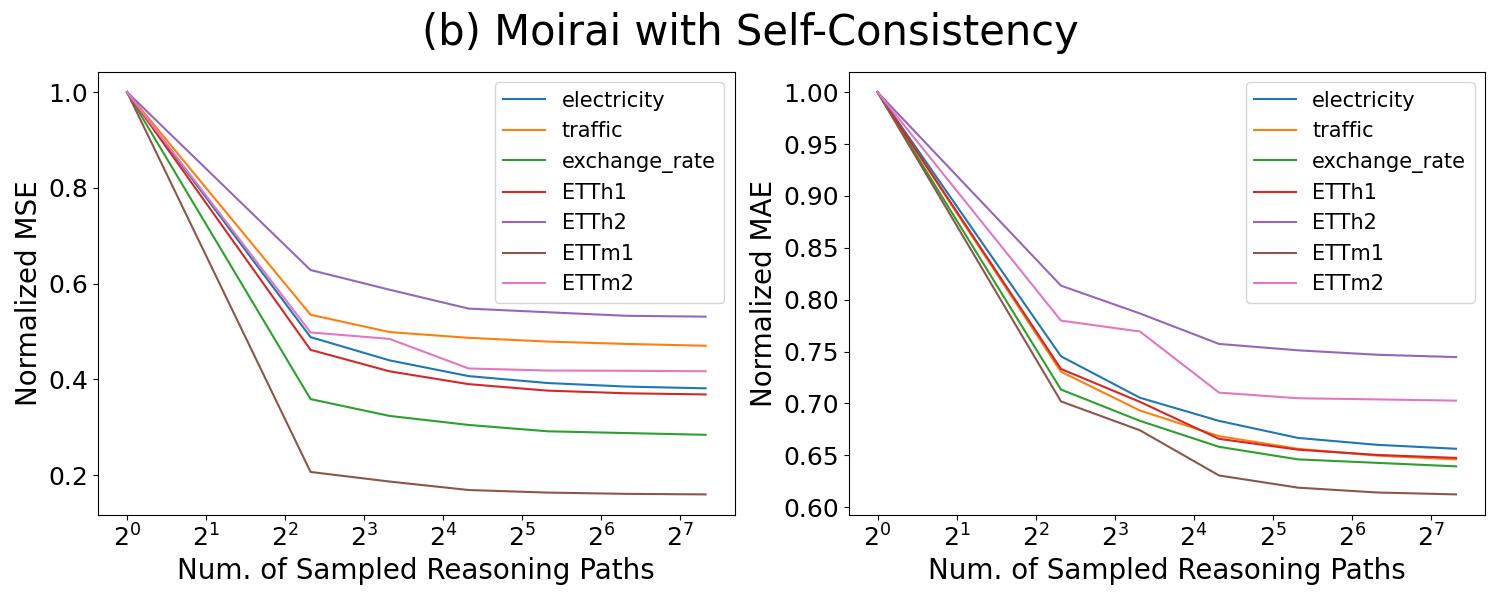}
    \caption{Verified test-time scaling law on foundation time-series models inspired by our insights. MSE and MAE are normalized based on the performance of one sampled path. The performance of Chronos and Moirai continuously improves as the number of sampled reasoning paths in the self-consistency reasoning strategy increases.}
    \label{fig:mse-exp}
\end{figure}

\par\noindent\textbf{Test-Time Scaling Law Verification.} We further generalize our empirical insights to foundation time-series models. Since the implementation of the GRPO-empowered reasoning model remains in the exploratory stage, we only focus on the self-consistency reasoning strategy. Specifically, we treat the sampling number of probabilistic foundation time-series models, including Moirai~\cite{woo2024unifiedtraininguniversaltime} and Chronos~\cite{ansari2024chronos}, as the number of sampled reasoning paths in self-consistency, and we use the median as the most consistent reasoning path. We use multiple well-adopted unimodal time-series datasets~\cite{wutimesnet}, and more experimental settings are detailed in Section~\ref{sec:SCALE}. As shown in Figure~\ref{fig:mse-exp}, we clearly demonstrate a new and simple test-time scaling law for TSF. We observe that with an increase in the number of sampled reasoning paths at test time, the performance of both foundation TSF models improves consistently and significantly, reducing the MSE error by at least 20\% and up to 50\%. We also observe that the model performance gradually converges at about 32 sampled reasoning paths. Our verified scaling law provides promising evidence and improvement room for reasoning-empowered TSF.

\vspace{0.05in}

% \subsection{Reasoning-Annotated TSF Datasets}
\par\noindent\textbf{Reasoning-Annotated TSF Datasets}
Inspired by recent research in post-training large reasoning models~\cite{zhou2023lima,wang2024deep,muennighoff2025s1}, we realized that another obstacle for foundation TSF reasoning models research, aside from evaluation suites and scaling laws, is the lack of a reasoning-annotated dataset. To this end, we curate the first reasoning-annotated TSF dataset, named \data, which pairs TSF task queries and answers with reasoning traces. Specifically, we adopt six advanced and diverse foundation models, including GPT-4o, o1-mini, Gemini-2.0-flash, Gemini-2.0-flash-thinking, DeepSeek-V3, and DeepSeek-R1, and record both the visible final output and the intermediate reasoning chain (if available) for TSF tasks. We set an appropriate temperature for each model and repeat the sampling, covering all datasets and settings in \method.  
We present more details and demos with reasoning trajectories in Appendix~\ref{data_detail}.

%% file: Section/Conclusion.tex
\section{Limitations and Future Work}
Since reasoning strategies for foundational time-series models have not yet been studied and are difficult to implement directly, we have to reuse foundational language models to explore effective TSF reasoning strategies. We envision that our benchmark and insights offer promising potential for future research on understanding and designing effective reasoning strategies for zero-shot TSF. First, our results show that only limited existing System 2 approaches benefit zero-shot TSF. This can be interpreted as a distinction in reasoning logic between time-series analysis or zero-shot TSF, and other types of reasoning that are inherently learned by most System 2 models. This finding encourages future studies to design reasoning strategies more tailored to time-series data and zero-shot TSF tasks. Second, in validating the effectiveness of reasoning in zero-shot TSF, our open-source toolkit—comprising both \method and \data—provides guidelines for future research. Specifically, we validate the scaling law of zero-shot TSF by incorporating reasoning within \method, paving the way for future work on large-scale foundation TSF models with reasoning capabilities. Additionally, we provide \data, the first TSF dataset with reasoning annotations, enabling deeper investigation and understanding of reasoning strategies in zero-shot TSF.

\section{Conclusion}
In this work, we introduce \method, the first comprehensive benchmark to evaluate the effectiveness of existing reasoning strategies in zero-shot TSF tasks. Through \method’s evaluations, we address two key questions: whether (RQ1) reasoning benefits zero-shot TSF, and what (RQ2) reasoning strategy is most effective. Our findings demonstrate that zero-shot TSF can indeed benefit from reasoning, and moreover, further demonstrate that self-consistency reasoning provides the most significant advantage in zero-shot TSF.
\section*{Acknowledgements}
This paper was supported in part by the NSF (Expeditions CCF-1918770, CAREER IIS-2028586, Medium IIS-1955883, Medium IIS-2106961, Medium IIS-2403240, PIPP CCF-2200269), CDC MInD program, Meta faculty gift, and funds/computing resources from Georgia Tech and GTRI.

%% file: Section/Appendix.tex
\newpage
\section*{Appendix}
\section{More Details of \method}\label{sec:detail_bench}
\subsection{Details of Datasets}
\begin{figure*}[ht]
    \centering
    \subfloat[Agriculture]{\includegraphics[height=0.1 \linewidth]{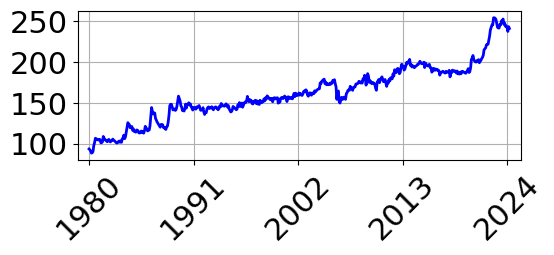}}
    \subfloat[Climate]{\includegraphics[height=0.1 \linewidth]{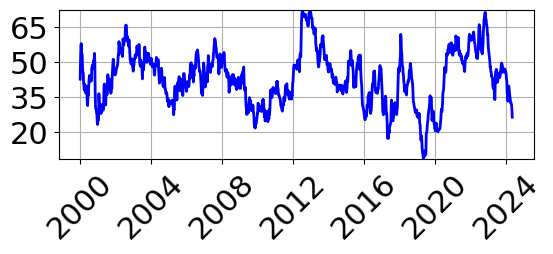}}
    \subfloat[Economy]{\includegraphics[height=0.1 \linewidth]{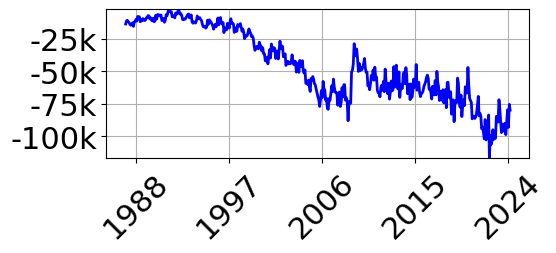}}
    \subfloat[Energy]{\includegraphics[height=0.1 \linewidth]{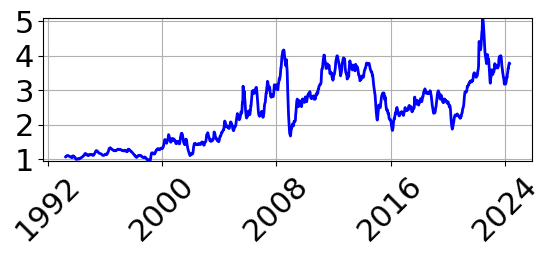}}
    \\
    \subfloat[Flu]{\includegraphics[height=0.12 \linewidth]{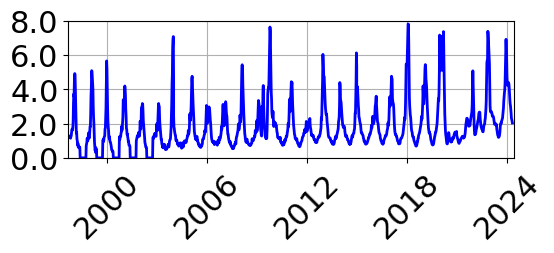}}
    \subfloat[Security]{\includegraphics[height=0.1 \linewidth]{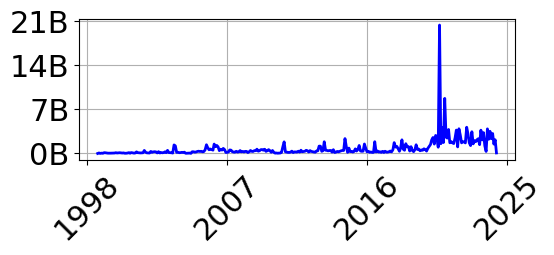}}
    \subfloat[Employment]{\includegraphics[height=0.1 \linewidth]{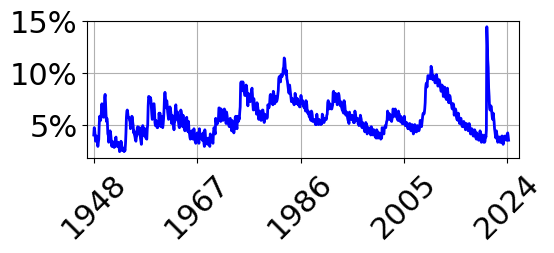}}
    \subfloat[Traffic]{\includegraphics[height=0.1 \linewidth]{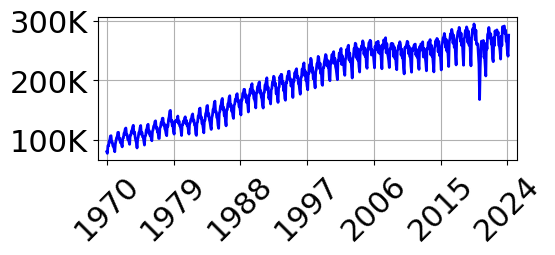}}
    \caption{Visualization of time-sereis data.}
    \label{fig:OT_visualize}
\end{figure*}

We provide details of datasets in Table~\ref{tab:series} and visualization in Figure~\ref{fig:OT_visualize}.
\begin{table*}[ht]
\centering
\caption{Overview of dataset in \method, covering key variables across nine domains with weekly, or monthly frequencies\label{tab:series}}
\resizebox{0.95\textwidth}{!}{

\begin{tabular}{ccccc}
\toprule
Domain & Target & Frequency & Timestamps & Timespan \\ \midrule
Agriculture & Retail Broiler Composite    & Monthly & 496496 & 1983 - 2024/04 \\
Climate  & Drought Level & Monthly & 496496 & 1983 - 2024/04 \\
Economy & International Trade Balance & Monthly & 423423 & 1989 - 2024/04 \\
Energy & Gasoline Prices & Weekly & 14791479 & 1996 - 2024/04 \\
Health & Influenza Patients Proportion & Weekly & 13891389 & 1997 - 2024/04 \\
Security & Disaster and Emergency Grants & Monthly & 297297 & 1999 - 2024/04 \\
Employment & Unemployment Rate & Monthly & 900900 & 1950 - 2024/04 \\ 
Traffic & Travel Volulmn & Monthly & 531531 &1980 - 2024/04 \\ \bottomrule
\end{tabular}
\vspace{-3mm}
}
\end{table*}
\subsection{Details of Reasoning Strategies}
\subsubsection{Implementation of reusing LLMs for unimodal forecasting}
Prompts used for unimodal time-series forecasting are detailed in Figure~\ref{fig:tsf_prompt}
\subsubsection{Implementation of reusing LLMs for multimodal forecasting}
Prompts used for multimodal time-series forecasting are detailed in Figure~\ref{fig:tsf_prompt_context}
\subsubsection{Implementation of chain-of-thought reasoning}
Prompts used for chain-of-thought reasoning are detailed in Figure~\ref{fig:tsf_prompt_cot}
\subsubsection{Implementation of self-consistency reasoning}
We set the default number of sampled reasoning paths in self-consistency to 3, using the median as the most consistent reasoning path. Prompts used for self-consistency reasoning are detailed in Figure~\ref{fig:self_consistency}
\subsubsection{Implementation of self-correction reasoning}
We set the default number of iterations for self-correction to 2. Prompts used for self-correction reasoning are detailed in Figure~\ref{fig:self_correction_code}
\begin{figure*}[ht]
    \centering
    \begin{tcolorbox}
    \begin{lstlisting}[breaklines=true]
base_prompt = f"""As an expert in {target_name}, predict the trends from {future_dates[0]} to {future_dates[-1]} 
based on both historical data and contextual information.

Historical data (chronological order):
{historical_str}

Contextual information:
{context_text}

Please enclose your final predictions between [PRED_START] and [PRED_END] markers exactly like this:
[PRED_START]
2024-01-01: 123.45
2024-02-01: 124.56
[PRED_END]"""
    \end{lstlisting}
    \end{tcolorbox}
    \caption{Prompt used for multimodal time-series forecasting.}
    \label{fig:tsf_prompt_context}
\end{figure*}

\begin{figure*}[ht]
    \centering
    \begin{tcolorbox}
    \begin{lstlisting}[breaklines=true]
base_prompt = f"""As an expert in {target_name}, predict the trends from {future_dates[0]} to {future_dates[-1]} based on the historical data.

Historical data (chronological order):
{historical_str}

Please enclose your final predictions between [PRED_START] and [PRED_END] markers exactly like this:
[PRED_START]
2024-01-01: 123.45
2024-02-01: 124.56
[PRED_END]"""
    \end{lstlisting}
    \end{tcolorbox}
    \caption{Prompt used for unimodal time-series forecasting.}
    \label{fig:tsf_prompt}
\end{figure*}
\begin{figure*}[ht]
    \centering
    \begin{tcolorbox}
    \begin{lstlisting}[breaklines=true]
prompt = base_prompt + "\n\nLet's approach this step by step:\n1. Analyze historical trends\n2. Identify patterns\n3. Make predictions"
    \end{lstlisting}
    \end{tcolorbox}
    \caption{Prompt used for chain-of-thought reasoning.}
    \label{fig:tsf_prompt_cot}
\end{figure*}
\begin{figure*}[ht]
    \centering
    \begin{tcolorbox}
    \begin{lstlisting}[breaklines=true]
elif method == "self_consistency":
    predictions = []
    for _ in range(3):
        pred = get_single_prediction(base_prompt, future, target_name, api_key, model)
        predictions.append(pred)
    return average_predictions(predictions)
    \end{lstlisting}
    \end{tcolorbox}
    \caption{Prompt used for self-consistency reasoning strategy.}
    \label{fig:self_consistency}
\end{figure*}
\begin{figure*}[ht]
    \centering
    \begin{tcolorbox}
    \begin{lstlisting}[breaklines=true]
elif method == "self_correction":
    current_pred = get_single_prediction(base_prompt, future, target_name, api_key, model)
    for _ in range(2):
        correction_prompt = f"{base_prompt}\n\nPrevious prediction:\n{format_prediction(current_pred)}\n\nPlease review and improve the prediction."
        current_pred = get_single_prediction(correction_prompt, future, target_name, api_key, model)
    return current_pred
    \end{lstlisting}
    \end{tcolorbox}
    \caption{Prompt used for self-correction reasoning strategy.}
    \label{fig:self_correction_code}
\end{figure*}

\subsection{Details of Models}

\subsubsection{OpenAI Model Series}
We select GPT-4o and o1-mini as representatives of System 1 and System2 models from OpenAI. Released on Mar 13 2024, GPT-4o is an autoregressive multimodal model which integrates all inputs and outputs into a unified neural network, offering faster and more cost-effective performance than its predecessors.\footnote{https://openai.com/index/gpt-4o-system-card/} In comparison, o1-mini is trained with large-scale reinforcement learning to reason, and it shows better reasoning performance over GPT-4o. \footnote{https://openai.com/index/openai-o1-mini-advancing-cost-efficient-reasoning/}

\subsubsection{Google Model Series}
Taken from the latest Gemini models, Gemini 2.0 Flash and Gemini 2.0 Flash Thinking compose a pair of System 1 and 2 models. Based on the non-thinking version, Gemini 2.0 Flash Thinking is an experimental model designed to enhance reasoning capabilities by generating its own "thinking process" during responses. This self-articulation allows the model to break down complex tasks into smaller steps, leading to more robust and accurate reasoning outcomes.  \footnote{https://ai.google.dev/gemini-api/docs/thinking}.

\subsubsection{Deepseek Model Series}
DeepSeek-V3 is an open source 671B Mixture-of-Experts (MoE) language model. It employs Multi-head Latent Attention and DeepSeekMoE architectures for efficient inference and cost-effective training. DeepSeek-R1 surpasses V3 in reasoning tasks by incorporating cold-start data before reinforcement learning (RL), enhancing readability and performance. This approach distills reasoning patterns from larger models into smaller ones, resulting in superior reasoning capabilities compared to those developed through RL on smaller models. 
 \footnote{https://github.com/deepseek-ai/DeepSeek-V3} \footnote{https://github.com/deepseek-ai/DeepSeek-R1}

\subsection{Details of Evaluations}
We use the API from popular OpenRouter~\footnote{\url{https://openrouter.ai/}} platform to perform LLM inference. We strictly follow the recommended parameters for each LLM, i.e., median values from users on OpenRouter. We detail the LLM model versions and hyperparameter settings in Figure~\ref{fig:model_params_1}.
\begin{figure*}[ht]
    \centering
    \begin{tcolorbox}
    \begin{lstlisting}[breaklines=true]
model_params = {
    "openai/o1-mini-2024-09-12": {
        "top_p": 1,
        "temperature": 0.7,
        "frequency_penalty": 0,
        "presence_penalty": 0,
        "repetition_penalty": 1,
        "top_k": 0
    },
    "google/gemini-2.0-flash-thinking-exp:free": {
        "top_p": 1,
        "temperature": 0.8,
        "repetition_penalty": 1
    },
    "deepseek/deepseek-r1": {
        "top_p": 1,
        "temperature": 0.7,
        "frequency_penalty": 0,
        "presence_penalty": 0,
        "repetition_penalty": 1,
        "top_k": 0
    },
    "openai/gpt-4o-2024-05-13": {
        "top_p": 1,
        "temperature": 0.9,
        "frequency_penalty": 0,
        "presence_penalty": 0,
        "repetition_penalty": 1,
        "top_k": 0
    },
    "google/gemini-2.0-flash-exp:free": {
        "top_p": 1,
        "temperature": 0.7,
        "frequency_penalty": 0,
        "presence_penalty": 0,
        "repetition_penalty": 1,
        "top_k": 0
    },
    "deepseek/deepseek-chat": {
        "top_p": 1,
        "temperature": 0.9,
        "frequency_penalty": 0,
        "presence_penalty": 0,
        "repetition_penalty": 1,
        "top_k": 0
    }
}
    \end{lstlisting}
    \end{tcolorbox}
    \caption{Hyperparameter settings for different models.}
    \label{fig:model_params_1}
\end{figure*}

\section{Details of Test-Time Scaling Law Experiments}\label{sec:SCALE}
\subsection{Data Setup}
We adopted the TimesNet datasets as shown in Table \ref{tab:Datasets of Test-Time Scaling Law Experiments} for our experiments in this section. All datasets listed are open source datasets.

\begin{table*}[b]
\centering
\caption{Dataset for Test-Time Scaling Law Experiments}
\resizebox{0.5\textwidth}{!}{

\begin{tabular}{cccc}
\toprule
Name & Domain & Frequency & Timestamps  \\ \midrule
Electricity & Energy    & Hourly & 26305  \\
Traffic  & Traffic & Hourly & 17545  \\
Exchange Rate & Economy & Daily & 7589  \\
ETTh1 & Energy & Hourly & 17421  \\
ETTh2 & Energy & Hourly & 17421  \\
ETTm1 & Energy & 15 Min & 69681  \\ 
ETTm2 & Energy & 15 Min & 69681 \\ \bottomrule
\end{tabular}
\vspace{-3mm}
}
\label{tab:Datasets of Test-Time Scaling Law Experiments}

\end{table*}

\subsection{Model Setup}
We utilized two contemporary popular zero-shot TSF that supports distributional forecasting, Chronos \cite{ansari2024chronos} and Moirai \cite{woo2024unifiedtraininguniversaltime}, to carry out our investigation about relationship between self-consistency and model performance. Here we take the number of sampled paths as a reflection of self-consistency. We chose the "base" version in experiments for both models.

\subsection{Hyperparameters}
The context for forecasting is set to 512 and the prediction length is set to 96, representing a short-term forecasting for 3 months. The specific numbers of sampled paths are set to 1 as a control group, then 5, 10 and doubling itself to 160 for scaling up.

\subsection{Numerical Results}
In table \ref{tab:Numerical MSE Chronos},\ref{tab:Numerical MAE Chronos},\ref{tab:Numerical MSE Moirai},\ref{tab:Numerical MAE Moirai} we include original validation results for our test-time scaling law, besides normalized ones mentioned above.

\begin{table*}[b]
\centering
\caption{Numerical Results of Chronos (MSE)}
\resizebox{0.95\textwidth}{!}{

\begin{tabular}{cccccccc}
\toprule
Num Path $\backslash$ Dataset & Electricity & Traffic & Exchange Rate & ETTh1 & ETTh2 & ETTm1 & ETTm2 \\ \midrule
1 & 101164.19 & 2.17e-5 & 2.12e-3 & 13.95 & 50.00 & 9.62 & 37.42 \\
5  & 63658.19 & 1.68e-5 & 1.14e-3 & 11.59 & 34.73 & 6.50 & 21.57  \\
10 & 59627.12 & 1.60e-5 & 1.13e-3 & 10.29 & 31.32 & 5.94 & 20.71  \\
20 & 55347.09 & 1.58e-5 & 1.02e-3 & 10.06 & 30.47 & 5.58 & 19.68  \\
40 & 54779.91 & 1.55e-5 & 1.00e-3 & 9.98 & 29.96 & 5.60 & 18.86  \\
80 & 54301.96 & 1.53e-5 & 1.00e-3 & 9.80 & 29.25 & 5.52 & 18.44 \\
160 & 54233.27 & 1.53e-5 & 0.90e-3 & 9.98 & 29.52 & 5.50 & 18.54 \\  \bottomrule
\end{tabular}
\vspace{-3mm}
}
\label{tab:Numerical MSE Chronos}

\end{table*}

\begin{table*}[b]
\centering
\caption{Numerical Results of Chronos (MAE)}
\resizebox{0.95\textwidth}{!}{

\begin{tabular}{cccccccc}
\toprule
Num Path $\backslash$ Dataset & Electricity & Traffic & Exchange Rate & ETTh1 & ETTh2 & ETTm1 & ETTm2 \\ \midrule
1 & 225.51 & 2.77e-3 & 0.033 & 2.77 & 5.12 & 2.23 & 4.33 \\
5  & 176.63 & 2.22e-3 & 0.024 & 2.51 & 4.31 & 1.83 & 3.40  \\
10 & 168.59 & 2.17e-3 & 0.023 & 2.38 & 4.11 & 1.76 & 3.32 \\
20 & 162.45 & 2.11e-3 & 0.022 & 2.35 & 4.05 & 1.69 & 3.20 \\
40 & 161.03 & 2.07e-3 & 0.022 & 2.33 & 4.05 & 1.69 & 3.13 \\
80 & 160.15 & 2.03e-3 & 0.022 & 2.29 & 3.97 & 1.67 & 3.11 \\
160 & 159.89 & 2.03e-3 & 0.022 & 2.32 & 3.98 & 1.66 & 3.11 \\  \bottomrule
\end{tabular}
\vspace{-3mm}
}
\label{tab:Numerical MAE Chronos}

\end{table*}

\begin{table*}[b]
\centering
\caption{Numerical Results of Moirai (MSE)}
\resizebox{0.95\textwidth}{!}{

\begin{tabular}{cccccccc}
\toprule
Num Path $\backslash$ Dataset & Electricity & Traffic & Exchange Rate & ETTh1 & ETTh2 & ETTm1 & ETTm2 \\ \midrule
1 & 181007.83 & 3.23e-5 & 3.80e-3 & 31.55 & 68.30 & 40.91 & 65.14 \\
5  & 88376.75 & 1.73e-5 & 1.36e-3 & 14.57 & 42.92 & 8.45 & 32.44 \\
10 & 79590.41 & 1.61e-5 & 1.23e-3 & 13.15 & 40.11 & 7.63 & 31.56 \\
20 & 73638.04 & 1.57e-5 & 1.16e-3 & 12.31 & 37.42 & 6.91 & 27.54 \\
40 & 71017.52 & 1.55e-5 & 1.11e-3 & 11.88 & 36.90 & 6.69 & 27.26 \\
80 & 69676.72 & 1.53e-5 & 1.09e-3 & 11.70 & 36.39 & 6.58 & 27.23 \\
160 & 69018.68 & 1.52e-5 & 1.08e-3 & 11.63 & 36.27 & 6.53 & 27.16 \\  \bottomrule
\end{tabular}
\vspace{-3mm}
}
\label{tab:Numerical MSE Moirai}

\end{table*}

\begin{table*}[b]
\centering
\caption{Numerical Results of Moirai (MAE)}
\resizebox{0.95\textwidth}{!}{

\begin{tabular}{cccccccc}
\toprule
Num Path $\backslash$ Dataset & Electricity & Traffic & Exchange Rate & ETTh1 & ETTh2 & ETTm1 & ETTm2 \\ \midrule
1 & 277.88 & 3.21e-3 & 0.036 & 3.78 & 5.95 & 2.97 & 5.45\\
5  & 207.14 & 2.34e-3 & 0.026 & 2.77 & 4.84 & 2.09 & 4.25 \\
10 & 196.03 & 2.23e-3 & 0.025 & 2.65 & 4.68 & 2.01 & 4.19 \\
20 & 189.85 & 2.15e-3 & 0.024 & 2.52 & 4.50 & 1.88 & 3.87 \\
40 & 185.28 & 2.11e-3 & 0.023 & 2.48 & 4.44 & 1.84 & 3.84 \\
80 & 183.44 & 2.09e-3 & 0.023 & 2.46 & 4.44 & 1.83 & 3.84 \\
160 & 182.38 & 2.07e-3 & 0.023 & 2.45 & 4.43 & 1.82 & 3.83 \\  \bottomrule
\end{tabular}
\vspace{-3mm}
}
\label{tab:Numerical MAE Moirai}

\end{table*}
\section{Details of Reasoning-Annotated TSF Datasets}\label{data_detail}
We use GPT-4o with the agriculture and climate dataset as an example to demonstrate the reasoning annotated TSF dataset we built. We organize the data in a Question and Answering format. 
\subsection{Agriculture}
\subsubsection{Unimodal TSF}
Demo is provided in Figure~\ref{fig:tsf_qa_example}.
\subsubsection{Multimodal TSF}
Demo is provided in Figure~\ref{fig:tsf_question_example}.
\subsection{Climate}
\subsubsection{Unimodal TSF}
Demo is provided in Figure~\ref{fig:tsf_drought_example}.
\subsubsection{Multimodal TSF}
Demo is provided in Figure~\ref{fig:tsf_drought_example_2}.
\begin{figure*}[ht]
    \centering
    \begin{tcolorbox}
    \begin{lstlisting}[breaklines=true]
## Question ##
"As an expert in the US Retail Broiler Composite, predict the trends from 2023-11-30 to 2024-01-31 based on the historical data.

Historical data (chronological order):
2015-11-30: 196.99; 2015-12-31: 193.15; 2016-01-31: 192.66; 2016-02-29: 187.42; ...
2023-08-31: 252.59; 2023-09-30: 246.16; 2023-10-31: 247.64

Please enclose your final predictions between [PRED_START] and [PRED_END] markers exactly like this:
[PRED_START]
2024-01-01: 123.45
2024-02-01: 124.56
[PRED_END]"

## Answer ##
"To make a prediction for the US Retail Broiler Composite prices for November 30, 2023, to January 31, 2024, let's first analyze the historical trends from previous years in these months.

Here's a summary of the data for November, December, and January over the years:
- Nov 2015: 196.99
- Dec 2015: 193.15
- Jan 2016: 192.66
...
- Nov 2022: 252.10
- Dec 2022: 247.03
- Jan 2023: 242.33

There appears to be a pattern where November values are not substantially different from December, and changes into January can vary. Recent years indicate a trend of increasing prices into December followed by a modest decrease in January.

For projection, considering recent trends showing an upward movement in prices, we'll apply a similar rate of change to the most current data points.

Given that the last available price is for October 2023 at 247.64, predicting a reasonable increase in November, a slight upward or stable trend in December, and then a small drop or stabilization in January:

[PRED_START]
2023-11-30: 249.75
2023-12-31: 250.85
2024-01-31: 248.50
[PRED_END]"
    \end{lstlisting}
    \end{tcolorbox}
    \caption{Reasoning annotated example of unimodal agriculture.}
    \label{fig:tsf_qa_example}
\end{figure*}
\begin{figure*}[ht]
    \centering
    \begin{tcolorbox}
    \begin{lstlisting}[breaklines=true]
## Question ##
"As an expert in the US Retail Broiler Composite, predict the trends from 2023-11-30 to 2024-01-31 
based on both historical data and contextual information.

Historical data (chronological order):
2015-11-30: 196.99; 2015-12-31: 193.15; 2016-01-31: 192.66; ...
2023-09-30: 246.16; 2023-10-31: 247.64

Contextual information:
2021-12-30: The United States broiler market has seen a 2% decrease in broiler-type chicks placed, ...
2023-10-26: The annual broiler per-capita consumption is reported in the Today's Beef Consumer Fall 2023 Update.

Please enclose your final predictions between [PRED_START] and [PRED_END] markers exactly like this:
[PRED_START]
2024-01-01: 123.45
2024-02-01: 124.56
[PRED_END]"

##Answer##
"Based on the historical data and contextual information provided, here is my prediction for the US Retail Broiler Composite from 2023-11-30 to 2024-01-31:\n\n1. **Historical Data Analysis**:\n   - Historically, prices have exhibited seasonal patterns with a tendency to decrease towards the end of the year and then either stabilize or slightly increase early in the year.\n   - November to January has shown some degree of variability, with occasional drops in prices in December followed by modest increases or stability in January.\n\n2. **Contextual Information**:\n   - The US broiler market faces various influencing factors like international demand, supply chain disruptions, labor shortages, and prices of feed like corn.\n   - The ongoing issues such as price-fixing charges among major poultry producers and bird flu have potential impacts on both supply and prices.\n   - The US market is connected to global trends, including export demands which are showing strong signals for growth in 2024.\n\n3. **Market Trends**:\n   - From recent reports, there is an expectation of a higher supply of chicken meat but with weaker domestic demand, potentially leading to slight price reductions.\n   - However, the international market remains robust, and overall inflation pressures could balance out some of the downward trends.\n\n4. **Conclusion**:\n   - Given the current market conditions, we can expect a slight decrease in prices towards the end of 2023 compared to the middle of the year. However, prices might stabilize or slightly increase in January 2024 as the market adjusts to new supply-demand dynamics and international pressures.\n\n[PRED_START]\n2023-11-30: 243.80\n2023-12-31: 241.50\n2024-01-31: 243.10\n[PRED_END]"
    \end{lstlisting}
    \end{tcolorbox}
    \caption{Reasoning annotated example of multimodal agriculture.}
    \label{fig:tsf_question_example}
\end{figure*}
\begin{figure*}[ht]
    \centering
    \begin{tcolorbox}
    \begin{lstlisting}[breaklines=true]
## Question ##
"As an expert in the US Drought Level, predict the trends from 2023-11-30 to 2024-01-31 
based on the historical data.

Historical data (chronological order):
2015-11-30: 3.35; 2015-12-31: 4.05; 2016-01-31: 2.04; ...
2023-09-30: 2.12; 2023-10-31: 2.17

Please enclose your final predictions between [PRED_START] and [PRED_END] markers exactly like this:
[PRED_START]
2024-01-01: 123.45
2024-02-01: 124.56
[PRED_END]"

## Answer ##
"To predict the US Drought Level trends from November 30, 2023, to January 31, 2024, 
we need to analyze historical seasonal patterns and recent trends.

### Seasonal Trends and Analysis:
1. **Late Fall to Early Winter (Nov to Jan):**
   - Historically, November to January sees diverse behavior but often a decrease from late fall to early winter.
   - Novembers have shown values like 1.75, 1.62, 2.71, 1.94, 1.94, 1.30, 2.44.
   - December and January values usually see some fluctuations, but generally, there's an inclination towards stabilizations or mild decreases.

2. **Recent Yearly Trends:**
   - 2022 and 2023 show end-of-year levels at:
     - 2022: 1.68 (Oct) to 2.44 (Nov to Dec) to 2.94 (Jan)
     - 2023: 2.17 (Oct latest data point)
   - We notice a slight rise in early winter periods, following potential weather patterns.

3. **Data Synthesis:**
   - Given slight upticks as we move into winter and the fact that the October 2023 value is 2.17, we can predict a modest rise by end of November.
   - Typical increases from November to January stay within a moderate range, respecting the patterns seen in similar past periods.

### Predictions:
Considering the above factors, here are the expected values:

[PRED_START]
2023-11-30: 2.35
2023-12-31: 2.55
2024-01-31: 2.75
[PRED_END]

This prediction reflects historical patterns, recent trends, and typical seasonal behaviors."
    \end{lstlisting}
    \end{tcolorbox}
    \caption{Reasoning annotated example of unimodal TSF for US Drought Level.}
    \label{fig:tsf_drought_example}
\end{figure*}
\begin{figure*}[ht]
    \centering
    \begin{tcolorbox}
    \begin{lstlisting}[breaklines=true]
## Question ##
"As an expert in the US Drought Level, predict the trends from 2023-11-30 to 2024-01-31 
based on both historical data and contextual information.

Historical data (chronological order):
2015-11-30: 3.35; 2015-12-31: 4.05; 2016-01-31: 2.04; ...
2023-09-30: 2.12; 2023-10-31: 2.17

Contextual information:
2021-12-30: Historical temperature, precipitation, and Palmer drought data from 1895 to present 
for climate divisions, states, and regions in the contiguous U.S. are available.
2023-10-26: The contiguous U.S. had its 18th-warmest October in the 129-year record.

Please enclose your final predictions between [PRED_START] and [PRED_END] markers exactly like this:
[PRED_START]
2024-01-01: 123.45
2024-02-01: 124.56
[PRED_END]"

## Answer ##
"Based on the historical data and contextual information provided, I will make predictions for 
the US Drought Level for the period from 2023-11-30 to 2024-01-31. I will take into account 
the trends in recent years and the seasonal patterns observed in the historical data, as well 
as the latest contextual information.

- Historical data suggests seasonal variations, with drought levels typically higher in the 
  winter months.
- Recent contextual data indicate an expected wetter-than-normal winter for the southern 
  contiguous U.S., which may alleviate drought conditions in those areas while potentially 
  leading to more pronounced drought in others.

Considering these factors, here are the predicted drought levels:

[PRED_START]
2023-11-30: 2.25
2023-12-31: 2.55
2024-01-31: 2.35
[PRED_END]"
    \end{lstlisting}
    \end{tcolorbox}
    \caption{Reasoning annotated example of multimodal TSF for US Drought Level.}
    \label{fig:tsf_drought_example_2}
\end{figure*}